\def\year{2021}\relax
\documentclass[letterpaper]{article} 
\usepackage{aaai21}  
\usepackage{times}  
\usepackage{helvet} 
\usepackage{courier}  
\usepackage[hyphens]{url}  
\usepackage{graphicx} 
\usepackage{natbib}  
\usepackage{caption} 
\usepackage{amsmath}
\usepackage[utf8]{inputenc}
\usepackage[english]{babel}
\usepackage{booktabs}
\usepackage{multirow}
\DeclareMathOperator*{\argmax}{arg\,max}

\usepackage{xcolor}
\usepackage{pifont}

\usepackage[linesnumbered,ruled]{algorithm2e}

\title{Knowledge Guided Learning: Towards Open Domain Egocentric Action Recognition with Zero Supervision}
\author {
        Sathyanarayanan N. Aakur,
        Sanjoy Kundu, 
        Nikhil Gunti \\
}
\affiliations {
     Department of Computer Science \\
    Oklahoma State University \\
    Stillwater, OK, USA 74078 \\
    \{saakurn, sanjoy.kundu, nikhil.gunti\}@okstate.edu
}
\begin{document}

\maketitle

\begin{abstract}
Advances in deep learning have enabled the development of models that have exhibited a remarkable tendency to recognize and even localize actions in videos. 
However, they tend to experience errors when faced with scenes or examples beyond their initial training environment. Hence, they fail to adapt to new domains without significant retraining with large amounts of annotated data. 
In this paper, we propose to overcome these limitations by moving to an open-world setting by decoupling the ideas of recognition and reasoning. 
Building upon the compositional representation offered by Grenander's Pattern Theory formalism, we show that attention and commonsense knowledge can be used to enable the self-supervised discovery of novel actions in egocentric videos in an open-world setting, where data from the observed environment (the target domain) is \textit{open} i.e., the vocabulary is partially known and training examples (both labeled and unlabeled) are not available.
We show that our approach can infer and learn novel classes for open vocabulary classification in egocentric videos and novel object detection with \textit{zero supervision}. 
Extensive experiments show its competitive performance on two publicly available egocentric action recognition datasets (GTEA Gaze and GTEA Gaze+) under open-world conditions.
\end{abstract}


\section{Introduction}
Computer vision models have taken great strides in action recognition in egocentric videos~\cite{fathi2012learning,liu2011recognizing,singh2016first,Sudhakaran_2019_CVPR,zhang2017first} and action localization~\cite{aakur2020action,jain2015objects2action}. Success has largely been achieved with supervised models driven by deep learning approaches trained in an inductive learning setting to learn data-driven associations between input and a fixed set of classes. 
However, there appears to be an implicit closed world assumption in these approaches, i.e., they assume that all observed data is composed of a static, known set of objects (nouns), actions (verbs), and activities (noun+verb combination) that are in 1:1 correspondence with the vocabulary from the training data.  
Hence, they fail to adapt to new domains without significant re-training with large amounts of annotated data.
We argue that this limitation stems from two common themes: inductive learning and knowledge representation. 
The former refers to their tendency to experience errors when faced with scenes or examples beyond their initial training environment and hence fail to adapt to new or similar domains. 
The latter refers to the need for significant, carefully curated re-training of existing models to learn new concepts. 
Hence, one must account for \textit{every eventuality} when training these systems to ensure their performance in real-world environments. The combination of these two issues means that current systems are restricted to narrow, domain-specific environments with specific, pre-defined rules.

\begin{figure}
    \centering
    \includegraphics[width=0.95\columnwidth]{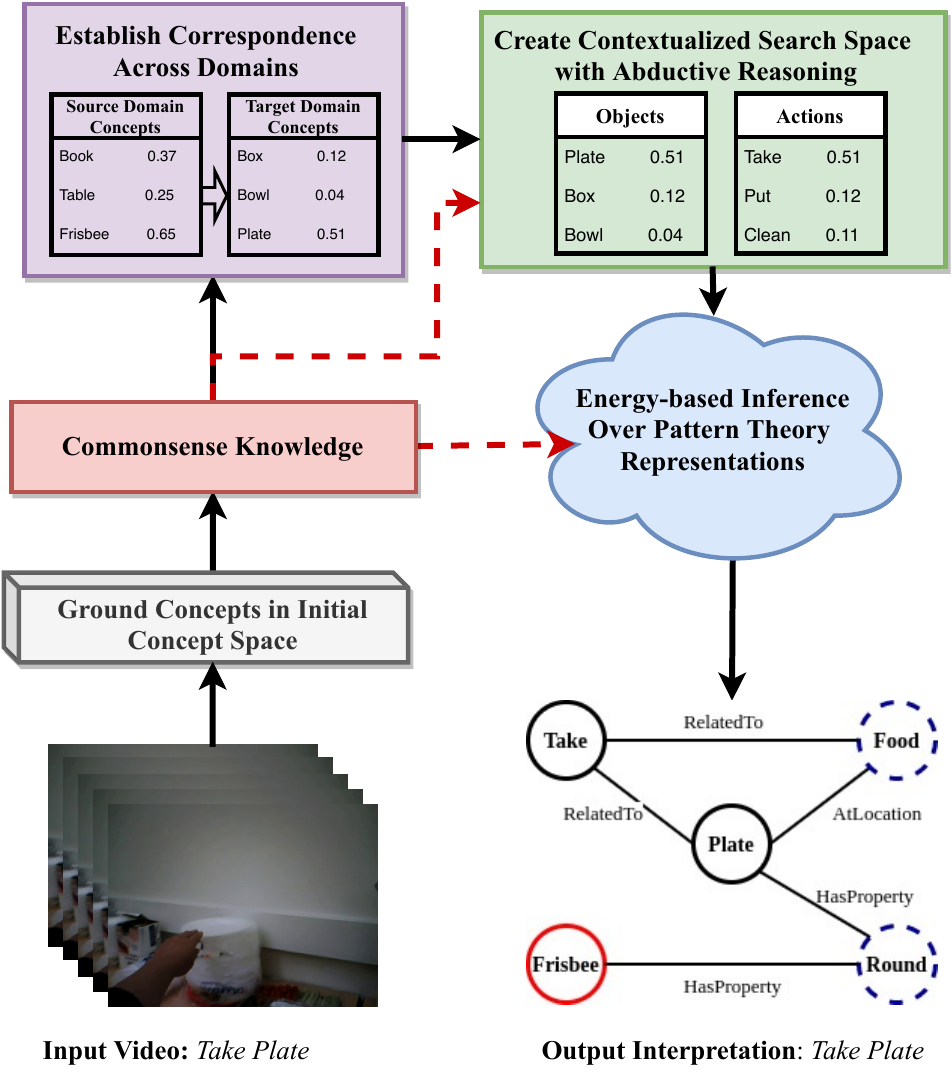}
    \caption{The proposed framework for open world inference in egocentric videos with zero supervision using commonsense knowledge.}
    \label{fig:arch}
\end{figure}

In this paper, we propose to move towards a more \textit{open-world} setting by augmenting the learning process with prior knowledge from largescale knowledgebases such as ConceptNet~\cite{liu2004conceptnet,speer2016conceptnet} and decoupling the ideas of recognition and reasoning. 
We leverage the compositional representation offered by Grenander's Canonical Pattern Theory formalism~\cite{grenander1996elements} and show that attention and commonsense knowledge can be used to enable the self-supervised discovery of novel actions in egocentric videos in an open-world setting. 
We show that our approach can be applied directly to open vocabulary classification in egocentric videos and show that it performs competitively with fully supervised and zero-shot learning baselines on two publicly available datasets. To our knowledge, this paper presents the first work to address the problem of open-world action recognition in egocentric videos with zero supervision. 

\textbf{What is an Open World?}
We consider an environment to be a \textit{closed world} if it \textit{only} consists of concepts (i.e., objects and actions), which have a one-to-one correspondence annotated training data that is available to a model. This is the case for most works trained in a supervised setting~\cite{singh2016first,Sudhakaran_2019_CVPR,ma2016going,ryoo2015pooled} where labeled concepts are static across training and test phases. 
In an entirely \textit{open world}, there are no such restrictions on the vocabulary. Any combination of concepts can co-occur in a given scene, beyond often what is captured in curated training data. 
While there can exist varying levels of ``openness'' across worlds, we consider the scenario where the set of elementary concepts is fixed but captured in a \textit{very large} vocabulary. 
Elementary recognition or detection models do not exist for all concepts. 
Hence the models have access to possible concepts that can exist in a scene but may not have encountered them during training. 
Note that this is different from unsupervised domain adaptation~\cite{kundu2020towards} and zero-shot learning~\cite{liu2011recognizing,zhang2017first,jain2015objects2action}. 
In \textit{zero-shot learning}, there are \textit{seen} and \textit{unseen} classes that are determined during training, but there is one key difference. The vocabulary is still fixed under a zero-shot setting, i.e., there are a fixed set of \textit{unseen} classes, which are a combination of elementary concepts such as actions and objects. 
In unsupervised domain adaptation, there is an implicit assumption that a labeled, source dataset and the presence of an unlabeled, target dataset are available for both training and inference. 
Specifically, the target dataset is assumed to have a set of labels which is a superset of the source dataset label space. 

\textbf{Contributions.} We make the following contributions in this work: (i) we are among the first to address the problem of open-world activity interpretation in egocentric videos with \textit{zero human supervision} i.e., we do not require target domain data or associated training annotations, (ii) we formulate a novel approach that integrates commonsense knowledge and symbolic reasoning with the representation learning capabilities of deep neural networks to overcome the dependence on annotated training data, (iii) we show that using compositional concept representations with Pattern Theory can learn semantic correspondences across domains and tasks, and (iv) we show that the proposed approach can extend beyond egocentric videos to learn novel concepts grounded in commonsense knowledge. 

\subsection{Related Work}
\textbf{Fully supervised approaches} treat egocentric action recognition as a \textit{supervised, classification} problem and assign the semantics to the video in terms of labels. Much of the recent success in egocentric action recognition~\cite{singh2016first,Sudhakaran_2019_CVPR,ma2016going,ryoo2015pooled} has been through the use of deep neural networks such as two-stream approach models~\cite{ma2016going}, attention-based models~\cite{Sudhakaran_2019_CVPR} and cascaded feature learning approaches~\cite{ryoo2015pooled}.

\textbf{Zero-shot learning} models~\cite{liu2011recognizing,zhang2017first,jain2015objects2action} and domain adaptation models~\cite{kundu2020towards} do not require as much supervision and learn semantic correspondences that extend beyond training classes to unseen test classes. The common approaches are to either use an \textit{attribute space} or embedding space that captures the semantics of a scene and helps extend beyond the training label by exploiting the semantic correspondences across classes. However, the success of such models relies on the presence of ``seen'' training classes that allow it to establish \textit{semantic} correspondences to recognize a finite set of unseen actions. We are not restricted by this constraint since we exploit an object's \textit{compositionality} and functionality to move beyond a fixed vocabulary. 

\textbf{Knowledge-driven recognition} has not been extensively explored in the existing literature. The most related approaches to ours are prior Pattern Theory-based frameworks~\cite{aakur2019generating,aakur2019wacv}, which also use a hybrid, symbolic approach to make inferences about activities in videos. However, they require access to the groundtruth annotations to train action (verb) and object (noun) detectors for \textit{each domain}. We do not have that limitation since we consider a \textit{more open world} where there is no requirement for obtaining training labels for elementary concepts. VideoBert~\cite{sun2019videobert} is another approach similar in spirit to ours but still requires large amounts of cross-modal data to pre-train associations to make inferences across domains. 

\section{Background: Building Compositional Representations}\label{sec:BG_Intro}
In this section, we \textit{briefly} introduce the Grenander's Pattern Theory formalism~\cite{grenander1996elements,aakur2019generating} for building compositional concept structures. 
We represent knowledge elements (or concepts such as actions and objects) that can be found in an environment through the flexible representation offered by Grenander's Pattern Theory formalism~\cite{grenander1996elements,aakur2019generating}. 
Each concept is represented by the basic, atomic unit of representation called a \textit{generator} $g_i\in G_s$, where $G_s$ is called the \textit{generator space}. Each generator represents the presence of a concept in a given observation. The generator space ($G_s$) is a finite collection of all possible concepts required to describe a scene. 
Hence, it can be divided into non-disjoint subsets $G^{\alpha}$, each representing a finite set of concepts that can exist in a given domain. For example, $G^{Kitchen}$ can represent the set of concepts (knife, spoon, cut, bake, batter, etc.) that are present in the kitchen domain, whereas $G^{zoo}$ represents the set of concepts that can exist in the \textit{zoo} domain. Hence the generator space is given by
$G_s = \bigcup_{\alpha\in A} G^{\alpha}$, where $A$ represents the set of all domains. 
In this work, we allow $A$ to be unconstrained, i.e., it can span all possible domains, and $\alpha$ is the \textit{generator index} that specifies a domain. 

Each generator has a set of links called \textit{bonds} that can be used to connect with other generators. Each generator $g_i$ has a fixed number of bonds called the \texttt{arity} of a generator denoted by $w(g_i) \forall g_i \in G_S$. Each bond represents a \textit{semantic assertion} that can be used to connect with other \textit{compatible generators} through bond interaction. Each bond is \textit{directed} and hence allows us to represent complex semantic structures that can capture the hierarchy of the semantic assertion being expressed. 
Bonds are quantified using the strength of the semantic relationships using the bond energy function:
\begin{equation}
e_{sem}(\beta^{\prime}(g_i),\beta^{\prime\prime}(g_j)) = \tanh(\phi(g_i,g_j)). 
\label{eqn:SemEnergy}
\end{equation}
where $\phi(.)$ represents the strength of the expressed assertion between concepts $g_i$ and $g_j$ through their respective bonds $\beta^{\prime}$ and $\beta^{\prime\prime}$. We use the semantic knowledge encoded in knowledgebases to populate the values of $\phi(.)$. 
$\tanh$ normalizes $\phi(.)$ between -1 to 1 to capture negative assertions between concepts.

\textbf{Expressing Complex Semantics.} Generators combine with other generators through compatible bonds to form complex structures called \textit{configurations}. Formally, a configuration $c$ is a connector graph $\sigma$ whose sites $1, \ldots, n$ are populated by a collection of generators $g_1, \ldots, g_n$ expressed as $c = \sigma (g_1, \ldots, g_i); g_i \in G_{S}$. The collection of generators $g_1, \ldots, g_i$ represents the semantic content of a given configuration $c$. We allow the connector graph to vary and hence define a set of all feasible connector graphs $\sigma$ to be $\Sigma$, known as the connection type. 
The probability of a configuration $c$ is a function of its total energy $E(c)$, which is defined as  
\begin{equation}
\begin{split}
E(c) &= -\sum_{ (\beta^{\prime},\beta^{\prime\prime})\in c}{e_{sem}(\beta^{\prime}(g_i),\beta^{\prime\prime}(g_j))}
\end{split}
\label{energy}
\end{equation}
and the probability of the configuration is given by $P(c) \propto  e^{-E(c)}$. Hence, lower energy indicates higher probability. 

\textbf{Knowledge Source: ConceptNet.} While our approach is general enough to handle multiple sources of commonsense knowledge such as OpenIE~\cite{martinez2018openie}, we use ConceptNet~\cite{liu2004conceptnet,speer2016conceptnet} as the source of general-purpose, commonsense knowledge to populate the generator space $G_s$, the bond structure of each generator and quantify semantic assertions ($\phi(\cdot)$). ConceptNet encodes cross-domain semantic information in a hypergraph, with nodes representing concepts connected through labeled edges, which specify and quantify the semantic relationship (or assertions) between concepts through edge weights.

\section{Open World Egocentric Action Recognition}\label{sec:actionRecog}
In this section, we introduce our open-world egocentric action recognition framework, as illustrated in Figure~\ref{fig:arch}. Our approach has four core components: (i) object-centric spatial region proposal, (ii) an attention-driven localization process, (iii) concept generator population, and (iv) an inference process to identify the current action being performed. 

\subsection{Initial Concept Space: Object-centric Perception}
In our approach, we begin with an initial concept space initialized by a base, source domain $G^s$ from which we expand our vocabulary with abductive reasoning. 
The concepts found in the MS COCO dataset~\cite{lin2014microsoft} form our base source domain. We use off-the-shelf object detection model in Faster R-CNN~\cite{ren2015faster,wu2019detectron2} as initial correspondence functions $f^s_d$ and $f^s_k$ to learn associations between a given input data $I_t$ and concept generators $\underline{g}_i\in G^s$. The former learns to localize the concepts in space, and the latter learns to associate concept labels with localization. We create region proposals ($r_i\in R$) in each input frame $I_t$ of a given video segment using these functions. We set the detection threshold to a relatively lower value ($\approx 0.25$) to account for any uncertainties that can arise when encountering novel scenes. We use these object-centric region proposals as plausible regions of attention to identifying the action.

\subsection{Selecting Concepts with Attention}
In arriving at a representation, we face a {\it packaging problem}~\cite{maguire2008speaking} i.e., given a novel visual scene and the \textit{many} observed objects, which should be chosen to understand the current action?
Following cognitive theories of attention~\cite{horstmann2016novelty}, we use the \textit{gaze} or the attention of the person performing the task to select regions that are most relevant to the current action. We use an energy-based selection algorithm that filters region proposals and returns the collection of proposals that have the highest probability of capturing the current action context. 
Energy is assigned to each bounding box ($r_i\in R$) at time $t$, and the top $k$ bounding boxes with the least energy are taken as the final proposals. The energy of a bounding box $r_i$ is defined as 
\begin{equation}
E(r_{i}) = w_\alpha\ \phi(\alpha_{ij}, {r}_{i}) + w_t \delta({r}_{i}, \hat{r}_{j}\})
\label{eqn:BB_Energy}
\end{equation}
where $\phi(\cdot)$ is a function that returns a value characteristic of the distance between the bounding box center and gaze location, $\delta(\cdot)$ is a function that returns the minimum spatial distance between the current bounding box and the closest bounding box from the previous time step $\hat{r}_{j}$.
The constants $w_\alpha$ and $w_t$ are scaling factors, and $\delta(\cdot)$ is used to enforce temporal consistency. 
In our experiments we set $w_\alpha = 0.75, w_t = 1.0$ and use $k=10$ bounding boxes from the previous time step.

\subsection{Constructing a Contextualized Generator Space}
Given the selected regions $r_i$ that have the maximum probability of association with the current action, the next step is to construct a contextualized generator space that allows us to transition from the COCO vocabulary $G^s$ to the target vocabulary $G^t$. First, we define a \textit{semantic mapping function} $f^t_k$ that returns the probability of the target object concepts $\hat{g}_j\in G^t$. 
The semantic mapping function is designed to capture both the \textit{semantic relatedness} and the \textit{contextualized semantic similarity} of two concepts. 
The \textit{semantic relatedness} is a measure of the semantic relationship shared between two concepts. 
It is distinct from similarity, which is a more \textit{compositional} measure of the semantic relationship. 
Hence, the mapping function would balance the compositional properties of concepts when computing the semantic correspondences across domains without degenerating into simple pairwise relatedness factors. 

\textbf{Semantic Mapping between Domains.} We define the semantic mapping function \mbox{$f^t_k:G^s\longrightarrow G^t$} to be a \textit{multi-valued, bijective} function that associates each generator $g_i \in G^s$, the source domain,  with a set of plausible generators $\hat{g}_j\in G_t$ in the target (co-)domain. 
The mapping function $f^t_k$ is \textit{multi-valued} i.e. it associates each generator $g_i\in G^s$ to one or more generators $\hat{g}_j\in G^t$. 
It is also a \textit{bijective function} i.e. $f^t_k$ is both \textit{injective} ($f^t_k(g_i) = f^t_k(g_j) \text{ iff. } g_i=g_j$) and \textit{surjective} (i.e. there exists function $\hat{g}_j=f^t_k(g_i) \forall \hat{g_j}\in G^t$, where $g_i\in G^s$). Hence, each concept generator in the target domain has at least one corresponding function that provides provenance from the source domain. 
Formally, it is defined as
\begin{equation}
    f_k(g_i,\hat{g}_j) = \sigma\Bigl(\frac{d_{cs}(g_i, \hat{g}_j)}{d_{rel}(g_i, \hat{g}_j)}\Bigr)
    \label{eqn:f_k}
\end{equation}
where $d_{cs}(\cdot)$ refers to the contextualized semantic similarity and $d_{rel}(\cdot)$ is the semantic relatedness between $g_i$ and $\hat{g}_j$, and $\sigma$ is a nonlinear function. We capture the compositional relationship between two concept generators by using the notion of contextualization~\cite{gumperz1992contextualization,aakur2019wacv,aakur2019generating}, which uses the relevant \textit{presuppositions} from prior knowledge to help establish semantic correspondences between concepts across domains. 
We compute the contextualized semantic similarity $d_{cs}(\cdot)$ as
\begin{equation}
    d_{cs}(g_i, \hat{g}_j) = \sum_{g_l,g_m\in P(g_i, \hat{g}_j)}^{n} \omega\phi(g_l,g_m)
    \label{eqn:c-ds}
\end{equation}
where $P(g_i, \hat{g}_j)$ is the shortest path between the concept generators $g_i$ and $\hat{g}_j$ in ConceptNet that has \textit{at least} one compositional assertion that connects them. We only consider the named assertions \textit{IsA} and \textit{HasProperty} to be compositional assertions. $\omega$ is a weight value that is used to penalize longer contextualization paths and hence mitigate the effect of noise and bias introduced due to the scale of ConceptNet. We sample $\omega$ from an exponential decay function characterized by the length of the path length between the two concepts and is defined as $\omega_k=\gamma(1-\epsilon)^{k}$, where $k$ is the distance from $g_i$ in the path $P(g_i, \hat{g}_j)$. The relatedness $d_{rel}(\cdot)$ exploits the analogical properties of ConceptNet Numberbatch~\cite{speer2016conceptnet} embedding and is computed by the cosine similarity between the embeddings. 

\begin{table*}[]
\centering
\caption{Object (noun) recognition performance in egocentric videos. We compare with fully supervised, zero-shot learning (ZSL), and an unsupervised random baselines along with variations of the proposed KGL approach with \textit{no access to any target domain data}.  $^*$ indicates leave-one-class-out evaluation.}
\begin{tabular}{|c|c|c|c|c|c|}
\toprule
\multirow{2}{*}{\textbf{Method}}  & \multirow{2}{*}{\textbf{Supervision}} & \textbf{Target Domain} & \textbf{Target Domain} & \textbf{GTEA Gaze} & \textbf{GTEA Gaze+}\\
 & & \textbf{Data} & \textbf{Annotations} & \textbf{(Accuracy)} &  \textbf{(Accuracy)}\\
\toprule
Two-stream CNN & Full & \ding{51} & \ding{51} & 38.05 & 61.87\\
IDT & Full & \ding{51} & \ding{51} & 45.07 & 53.45\\
Action Decomposition & Full & \ding{51} & \ding{51} & 60.01 & 65.62\\
Action Decomposition (ZSL)$^*$ & Full & \ding{51} & \ding{51} & 40.65 &  43.44\\
\midrule
Random (Chance) & None & \ding{55} & \ding{55} & 3.22 & 3.70\\
Object-based ZSL & None & \ding{55} & \ding{55} & 3.97 & 4.50\\
KGL (Top-1) & None & \ding{55} & \ding{55} & 5.12 & 14.78\\
KGL (Top-5) & None & \ding{55} & \ding{55} & 10.73 & 37.99\\
KGL (Top-10) & None & \ding{55} & \ding{55} & 21.15 & 56.84\\
\midrule
KGL w/o Gaze (Top-10) & None & \ding{55} & \ding{55} & 15.68 & 35.43 \\
KGL w/ Action Oracle (Top-10) & None & \ding{55} & \ding{55} & 32.61 & 63.15\\
\bottomrule
\end{tabular}
\label{tab:objRecog}
\end{table*}

\begin{algorithm}
    \SetKwInOut{Input}{Input}
    \SetKwInOut{Output}{Output}

    MCMC Simulated Annealing $(R, G, U, \alpha, p, k_{max}, T_0)$\;
    $c \leftarrow resetConfiguration(R,G,U)$\\
    $best \leftarrow c$\\
    \For {$k\leftarrow 1\ldots k_{max}$:}
        {
        	$ t \leftarrow UniformSample(0,1) $\\
            \If {$t < p$}
            {
	            $c^{\prime} \leftarrow globalProposal(R,G)$
            }
            \uElse
            {
            	$c^{\prime} \leftarrow localProposal(c, G, U)$
            }
            $T \leftarrow T_0 \times \alpha^k $\\
            \If{$E(c^{\prime}) < E(c)$}
            {
            	$c \leftarrow c^{\prime}$\\
            }
            \uElse
            {
            	$ z \leftarrow UniformSample(0,1) $\\
                \If{$z < exp(-(E(c^{\prime}) - E(c))/T)$}
                {
                	$c \leftarrow c^{\prime}$\\
                }
            }
            \If{$E(c) < E(b)$}
            {
            	$best \leftarrow c$\\
            }
        }
    \Return best
  \caption{Open-world activity inference.}
  \label{MCMC_Inference}
\end{algorithm}

\textbf{Building the Concept Space.} For a given concept $\hat{g}_j$ and a detected concept $\underline{g}_i$, we define its probability as 

\begin{equation}
    p(\hat{g}_j|\underline{g}_i,r_i) = \frac{p(\underline{g}_i|r_i, I_t) * f_k(\hat{g}_j, \underline{g}_i)}{E(r_i)}
    \label{eqn:object_prob}
\end{equation}
where $f_k$ is the contextualized semantic mapping function from Equation~\ref{eqn:f_k} and $I_t$ is the input frame at time $t$. This function allows us to quantify the probability of presence of a novel concept $\hat{g}_j$ in a given scene \textit{without any training data, both labeled and unlabeled}. Note that the above function only allows us to build a concept space involving objects i.e. \textit{nouns} since it is possible to define compositional relationships to establish \textit{similarity} across domains to ensure that the concepts can be used interchangeably, particularly with respect to their \textit{affordance}.

To generate \textit{action} concepts that could be associated with the constructed object concept space, we tackle this problem through the notion of \textit{abductive reasoning} through which we generate and evaluate multiple candidate hypotheses (action or verb concepts) in a given domain ($G^t$). This allows us to constrain the search space to those concepts that share the \textit{affordance} of the object concepts in the given domain. 
Formally, we define abductive reasoning to be an optimization process that aims to find the optimal action generator $\tilde{g}_i \in \{\tilde{g}_1, \tilde{g}_2, \tilde{g}_3, \ldots \tilde{g}_n\}$ that has the maximum affordance conditioned upon the observed object generators $g_k\in G^s\cup G^t$ and prior, commonsense knowledge, $C_t$. This can be expressed as the optimization for 
\begin{equation}
     \argmax_{\tilde{g}_i \in \{\tilde{g}_1, \tilde{g}_2, \tilde{g}_3, \ldots \tilde{g}_n\}} p(\tilde{g}_i | C_t, g_k)
    \label{eqn:AbductiveProb}
\end{equation}
where $g_k$ represents the observed object concept in the target domain $G^j$ and its corresponding concept from the source domain $G^i$. This optimization involves the empirical computation of the probability of occurrence for each action or verb hypothesis $\tilde{g}_i$ given the commonsense knowledge $C_t$, captured in ConceptNet.
To account for uncertainty, we use the top-$K$ object and action labels. The probability of each action concept $\hat{g}^a_j$ is a measure of the object confidences from both the source ($\underline{g}_i$) and target ($\hat{g}_j$) domains and is defined as 
\begin{equation}
    p(\hat{g}^a_j|\underline{g}_i,\hat{g}_j) = \frac{d_{rel}(\hat{g}^a_j, \underline{g}_i) \cdot d_{rel}(\hat{g}^a_j, \hat{g}_j)}{d_{cs}(\hat{g}_j, \underline{g}_i)}
    \label{eqn:action_prob}
\end{equation}
where $d_{rel}(\cdot)$ is the semantic similarity between two concepts given by the cosine similarity between the ConceptNet Numberbatch~\cite{speer2016conceptnet} vector embedding of the two concepts. Action probabilities are a function of object confidence and the semantic correspondence between concepts across domains. 
Candidate action concepts ($\tilde{G}^j=\{\tilde{g}_1, \tilde{g}_2, \tilde{g}_3, \ldots \tilde{g}_n\}$) can be pre-defined using domain knowledge or expanded through ConceptNet traversal using the contextualized similarity path $P(g_i, \hat{g}_j)$ ( Equation~\ref{eqn:c-ds}). The former is a closed world with a small search space while the latter is an open world with \textit{an unconstrained vocabulary}. 
We experiment with both and show that the proposed approach is a significant step towards \textit{completely} open-world learning with unconstrained vocabulary.

\subsection{Inference}
Given the putative object and action labels, we define an inference function that reasons about the semantic relationships between these concepts to arrive at an interpretation of the scene. 
Since we allow for multiple possibilities in both action and object space, a feasible optimization solution for such an exponentially large search space is a sampling strategy. 
We follow the work in~\cite{aakur2019wacv} and employ a Markov Chain Monte Carlo (MCMC) based simulated annealing process, which uses two proposal functions for inference. 
A global proposal function samples an underlying connector graph $\sigma$ for an interpretation, and the local proposal populates the sites in the connector graph. 
Each jump gives rise to a configuration whose semantic content represents a possible interpretation for the given video. 
Configurations with the least energy represent possible interpretations of the activity.
The algorithm for the MCMC-based simulated annealing process is shown in Algorithm~\ref{MCMC_Inference}. We begin with the set of \textit{filtered} region proposals with detected concepts from the source domain $G^s$, the corresponding set of plausible target domain generators $G$, and concept generators from ConceptNet $U$, that provide the background knowledge for concept generators. 
We initialize the search by sampling an initial configuration $c^\prime$. 
The proposed configuration is used as initialization for the ``\textit{best}'' configuration seen so far.
The search is initiated and performed for a fixed number of iterations $k_{max}$ defined in the parameters. The choice between the proposal functions is decided by sampling from a uniform distribution. 
At each step of the annealing process, the temperature is updated based on a cooling rate given by $\alpha^k$, where $\alpha$ is a predefined constant. 
Each step of the simulated annealing process yields a new configuration $c^\prime$, accepted or rejected, based on its energy. 
\section{Experimental Evaluation}
\begin{table*}[]
\centering
\caption{Action (verb) and activity (verb+noun) recognition performance in egocentric videos. We compare against supervised and zero-shot learning (ZSL) baselines along with variations of the proposed open-world KGL approach. $^*$ indicates performance reported for leave-one-class-out evaluation.}
\begin{tabular}{|c|c|c|c|c|c|c|c|c|c|}
\toprule
\multirow{2}{*}{\textbf{Method}}  & \multirow{2}{*}{\textbf{Supervision}} & \textbf{Target Domain} & \textbf{Target Domain} & \multicolumn{2}{c}{\textbf{GTEA Gaze}} & \multicolumn{2}{|c|}{\textbf{GTEA Gaze+}}\\
\cline{5-8}
 & & \textbf{Data} & \textbf{Annotations} & \textbf{Verb} & \textbf{Activity} &  \textbf{Verb} & \textbf{Activity}\\
\toprule
Two-stream CNN & Full & \ding{51} & \ding{51} & 59.54 & 53.08 & 58.65 & 44.89\\
IDT & Full & \ding{51} & \ding{51} & 75.55 & 40.41 & 66.74 & 51.26\\
Action Decomposition & Full & \ding{51} & \ding{51} & 79.39 & 55.67 & 75.07 & 57.79\\
Action Decomposition (ZSL)$^*$ & Full & \ding{51} & \ding{51} & 85.28 & 39.63 & 27.68 & 15.98\\
\midrule
Random (Chance) & None & \ding{55} & \ding{55} & 7.69 & 2.50 & 4.55 & 2.28  \\
Object-based ZSL & None & \ding{55} & \ding{55} & 6.45 & 3.81 & 5.69 & 6.58\\
KGL (Top-1) & None & \ding{55} & \ding{55} & 8.21 & 4.91 & 6.73 & 10.87\\
KGL (Top-5) & None & \ding{55} & \ding{55} & 32.39 & 18.78 & 24.64 & 27.53\\
KGL (Top-10) & None & \ding{55} & \ding{55} & 50.72 & 30.97 & 36.62 & 38.59\\
\midrule
KGL w/o Gaze (Top-10) & None & \ding{55} & \ding{55} & 33.56 & 22.67 & 27.54 & 26.21\\
KGL w/ Object Oracle (Top-10) & None & \ding{55} & \ding{55} & 53.89 & 36.47 & 40.95 &  50.78\\
\bottomrule
\end{tabular}
\label{tab:verbAcc}
\end{table*}
\textbf{Data.} We use the GTEA Gaze~\cite{fathi2012learning} and the GTEA Gaze+~\cite{li2013learning} as our test environment for open-world object, action, and activity recognition in egocentric videos. The two datasets consist of several video sequences on meal preparation tasks by different subjects and ground-truth annotations of their gaze positions. 
The activity annotations consist of an action (verb) and the corresponding object (noun). GTEA Gaze contains 10 different verbs and 38 different nouns, while GTEA Gaze+ contains 15 verbs and 27 nouns. 
We also test our approach's generalization capability to scenes beyond egocentric videos for generalized object detection. We use a subset of Open Images~\cite{kuznetsova2018open} with 10 classes called the \textit{Open Images OW-10} dataset with $3095$ images and $5686$ bounding box annotations.
Each of these 10 classes can be found in the GTEA Gaze dataset. It allows us to evaluate our model beyond egocentric videos where the gaze positions isolate the object of interest. 
The goal is to expand the vocabulary beyond MS COCO \textit{without any supervision}. 

\textbf{Metrics.} We report the accuracy for action (verb) and object (noun) recognition. For activity recognition (i.e., verb+noun prediction), we use the concept of word accuracy in speech~\cite{kuehne2014language} to measure the semantic similarity between the prediction and ground-truth. 
We report the recall per $100$ predictions and the mean average precision at $0.5$ overlap and the mean over $0.5:0.95$ thresholds for object detection. Note that we report accuracy for top-k predictions ($k \in\{1,5,10\}$) for the open-world KGL approaches, since the prediction is under an \textit{open-world activity recognition setting}. We do not predict nouns and verbs independently but make a joint inference over the activity (verb+noun) classes. The total number of verbs and nouns in GTEA is 10 and 38, respectively, resulting in 380 noun+verb combinations. Similarly, the number for GTEA+ is 15 verbs, 27 nouns, and 405 noun+verb combinations. Hence top-k accuracy over this search space is considered and not over nouns/verbs individually. This is not an unreasonable evaluation setting considering that the search space for verb-noun pairs is quite large and can grow exponentially as the number of the plausible nouns and verb candidates increase. 

\textbf{Baselines.} We establish baseline approaches with different supervision needs. We employ a two-stream CNN~\cite{ma2016going}, Improved Dense Trajectories (IDT)~\cite{wang2013action} and Action Decomposition~\cite{zhang2017first}. We consider two zero-shot learning (ZSL) baselines. First, we use the zero-shot variation of Action Decomposition~\cite{zhang2017first} as a standard baseline for ZSL egocentric action recognition, where the approach has access to the target domain except for one \textit{unseen} action. Second, we develop an alternative ZSL approach called ``\textit{Object-based ZSL}'', with a more open-world setting. The approach has the same initial concept space as our approach (MS COCO). It uses cosine similarity between ConceptNet Numberbatch embedding of object-level labels from object detection models and the target verb+noun combination to generate action labels during inference. 
For the unsupervised, open-world setting, we use three baselines: (i) a random prediction model, (ii) our approach without attention, and (iii) our approach with an oracle, i.e., perfect target domain object detection or action label. Note that the random prediction model predicts a random verb+noun activity per video and not individual actions and objects. 
Both the supervised and ZSL approaches have access to both the videos and annotations from the target domain, while our approach only has access to the set of \textit{noun} and \textit{verb} concepts in the target domain. 
We include two oracle models, object oracle and action oracle, to evaluate the performance of our model when some of the training annotations are known, hence reducing the search space. These two models are intended to evaluate a less open world setting, where there is an available object detection \textit{or} action recognition model whose labels are a super-set of the target environment’s label space. Given the increasing emphasis on large and diverse datasets, such as Google Open Images~\cite{kuznetsova2018open} dataset and Kinetics-800~\cite{carreira2017quo}, these \textit{oracle}-based settings are likely to become a very common setting in the future.

\subsection{Learning to Recognize Novel Objects}
We first evaluate our approach's ability to recognize objects across domains, i.e., transfer from MS COCO to GTEA Gaze or Gaze+ with zero supervision. This evaluation allows us to assess our semantic mapping function's ability to learn correspondences across tasks and domains. 
We summarize the results in Table~\ref{tab:objRecog} and compare them with both supervised and open-world baselines. It can be seen that our approach generalizes well across domains \textit{without any supervision, including access to target domain data}.
This is a key difference between our approach and other models (including ZSL approaches), which have access to the target domain data for other known classes and are only expected to learn correspondences for a small number of ``unseen'' classes. 
Note that our approach uses a general-purpose knowledge base that is not specifically tailored for the kitchen domain and has no learned correspondence in the target domain. 
However, the top-5 and top-10 accuracy metrics show that our model achieves comparable performance to both fully supervised and ZSL baselines, without using any supervision from the target domain, indicating that the model can perform competitively in an open-world setting, where there is no access to target data (both seen and unseen) and annotations.
These results indicate that symbolic knowledgebases can be leveraged to help generalize object and activity recognition across domains with limited supervision.

\subsection{Learning Novel Actions}
Finally, we evaluate the model's ability to perform \textit{reasoning} over the object's functionality to identify the action (verb) and the activity being performed in the scene. We evaluate on two settings: (i) when the possible actions are known and (ii) when the potential list of actions is unrestricted, i.e., a completely open world.
\begin{table}[t]
\centering
\caption{Action recognition performance with unknown action space. 
}
\resizebox{0.99\columnwidth}{!}{
\begin{tabular}{|c|c|c|c|c|}
\toprule
\multirow{1}{*}{\textbf{Top-K}} & \multicolumn{4}{c|}{\textbf{Verb Accuracy}} \\
\cline{2-5}
 \textbf{Candidates}& \textbf{Top-1 Obj} & \textbf{Top-5 Obj} & \textbf{Top-10 Obj} & \textbf{Random}\\
\toprule
10 & 0.3 & 1.3 & 2.7 & 0.1\\
25 & 2.4 & 2.9 & 12.9 & 0.04\\
50 & 7.4 & 8.4 & 21.54 & 0.02\\
\bottomrule
\end{tabular}
}
\label{tab:openWorldActPerf}
\end{table}
Table~\ref{tab:verbAcc} shows the performance of the model in the former setting. It can be seen that our approach performs competitively with supervised and zero-shot baselines. 
We achieve a top-1 performance of 8.21\% for verb recognition on GTEA Gaze and 6.73\% on GTEA Gaze+, but it is to be noted that the search space for verb-noun pairs is quite large and can grow exponentially as more descriptive labels are imposed (subject-verb-object, etc.). Supervised models and zero-shot learning models do not take this into account as they assume that the label space is known and tractable (100 actions for GTEA and 44 for GTEA+). We do not have access to this information during inference, yet obtain a reasonable prediction accuracy at top-1. Increasing the threshold to top-5 and top-10 increases the performance significantly, showing that the reasoning results in relevant vocabulary despite unknown verb-noun combinations and the resulting search space.
%
We also experiment with an unrestricted action (verb) space and summarize the results in Table~\ref{tab:openWorldActPerf}. We take a varying number of top-$K$ object labels and their respective top-$K$ plausible action labels generated by traversing the path $P(\cdot$ from Equation~\ref{eqn:c-ds}) and run inference to obtain an interpretation from these inputs. We report the accuracy for the top-25 interpretations since the task is challenging, and the search space is rather large. 
As can be seen from Table~\ref{tab:openWorldActPerf}, we obtain a verb accuracy of $21.54\%$ when we use top-10 object labels and top-50 action labels for inference. 
Furthermore, we find at least one action from the target vocabulary with an accuracy of $83.7\%$ in the top-25 action labels. Considering that the \textit{verb} vocabulary is completely unknown, these results represent a significant step towards open-world action recognition.

\subsection{Localizing Objects beyond Egocentric Videos}
We also evaluate our model to perform object detection beyond egocentric videos by evaluating on the Open Images OW-10 subset. We use a Faster R-CNN model trained on MS COCO to generate region proposals and use the predicted labels to establish semantic correspondences to the target domain labels. We summarize the results in Table~\ref{tab:openImgsPerf}. It can be seen that we make a considerable improvement over a random baseline which predicts a random class for each bounding box. It should be noted that the classes do not have any overlap with MSCOCO and hence constitute a set of objects \textit{that have not been seen by the detector}. 
Additionally, the data from Open Images were not used in the training stage at any point. 
This differentiates us from zero-shot learning approaches which have access to the target domain data and partial access to groundtruth. This allows them to exploit context and feature-based similarity measures to recognize/detect unseen classes. We do not have access to any of the data and hence provides a completely novel environment where there are novel, unseen objects present and hence allows us to quantify the object detection capabilities of our approach. 
\begin{table}[t]
\centering
\caption{Open World Object detection on Open Images OW-10 Dataset. 
}
\resizebox{0.85\columnwidth}{!}{
\begin{tabular}{|c|c|c|c|}
\toprule

\multirow{2}{*}{\textbf{Method}}            & \textbf{mAP} & \textbf{mAP } & \multirow{2}{*}{\textbf{Recall}} \\
& \textbf{IOU=0.5} & \textbf{IOU=0.5:0.95} &  \\
\toprule
Ours 1 pred/BB & 0.8     & 0.5          & 5.9    \\
Ours 3 pred/BB & 10.2    & 8.0          & 24.6   \\
Ours 5 pred/BB & 8.1     & 6.2          & 35.5   \\
\midrule
Random & 0.1 & 0.05 & 2.7\\
\bottomrule
\end{tabular}
}
\label{tab:openImgsPerf}
\end{table}
We allow for multiple label predictions per bounding box proposal to account for uncertainty. It can be seen that the mAP at $0.5$ IOU and the mean over IOU ranges from 0.5:0.95 are remarkable, considering that no training data was used. It is interesting to note that using each region proposal for more than one object label results in better performance (mAP of $10.2$, IOU=$0.5$) but tapers off considerably when many labels are considered per proposal. 
We find that generating predictions using top-3 labels per proposal has the best performance, while using top-5 labels performs worse. 
We attribute it to the fact that the confidences become diluted when adding labels obtained through semantic correspondences.

\subsection{Qualitative Discussion}
\begin{figure*}
    \centering
    \begin{tabular}{ccc}
        \includegraphics[width=0.25\textwidth]{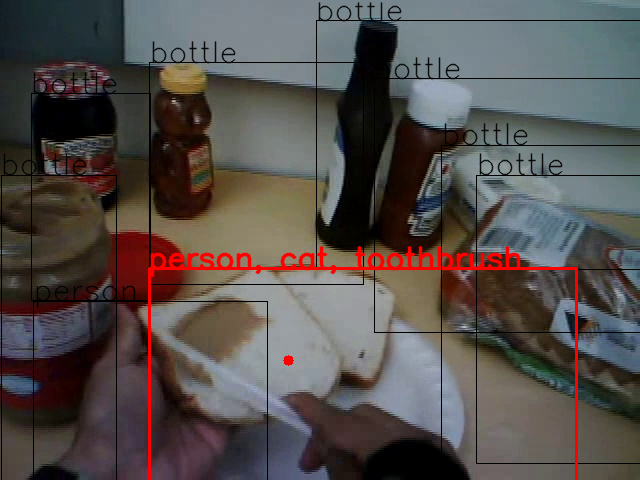} & \includegraphics[width=0.25\textwidth]{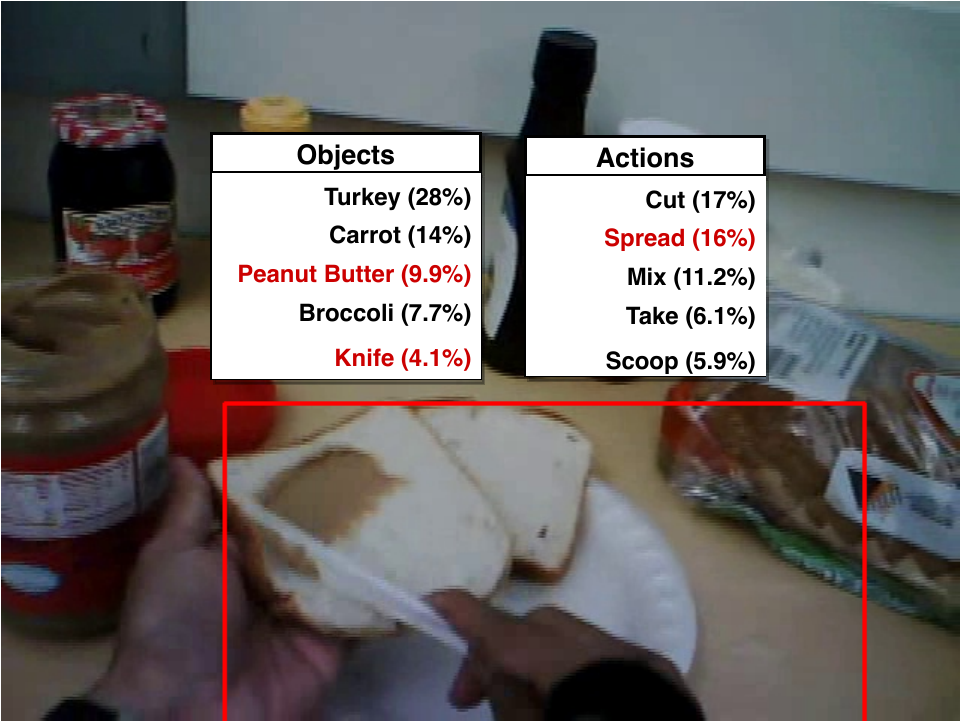} & \includegraphics[width=0.21\textwidth]{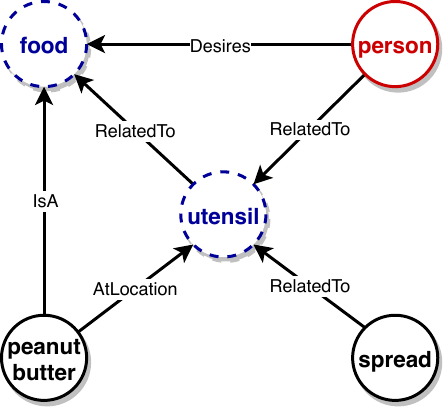}\\   
        \multicolumn{3}{c}{(a)}\\
        \includegraphics[width=0.25\textwidth]{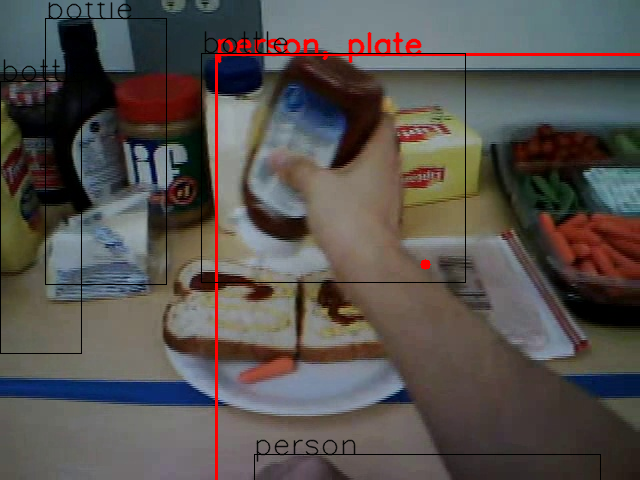} & \includegraphics[width=0.25\textwidth]{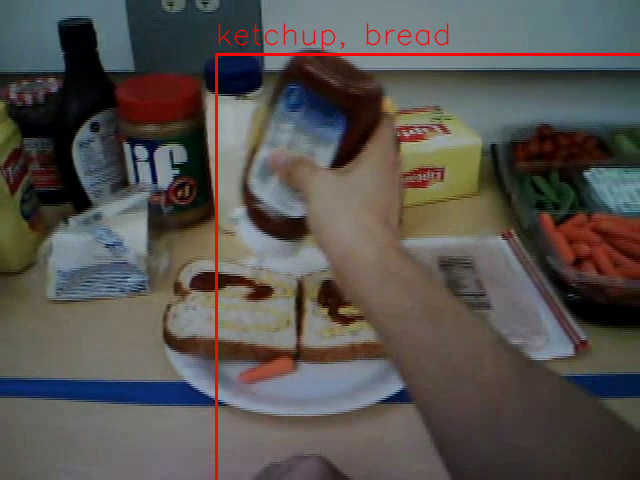} & \includegraphics[width=0.21\textwidth]{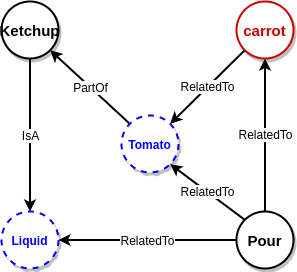}\\   
        \multicolumn{3}{c}{(b)}\\
    \end{tabular}
    
    \caption{    Visualization of the reasoning process at different stages when given a video, (a) ``\textit{spread peanut butter}'' and (b) ``\textit{pour ketchup}''. The first column visualizes the scene in the initial vocabulary, the second after establishing semantic correspondence, and the last shows the final interpretation.
    }
    \label{fig:qualViz}
\end{figure*}
An interesting property of our approach is that it can handle uncertainty in elementary concept recognition and is not restricted to the vocabulary of the training annotations.
Supervised and zero-shot models have a more restricted vocabulary that constrains the vocabulary to action-object combinations seen in the training annotations. We show an example in Figure~\ref{fig:qualViz}, where our model was able to arrive at the correct interpretation even when the target noun and verb were not the top-$1$ prediction. 
We illustrate the contextualized path (dotted circles) for completeness. 
The simulated annealing-based inference allows for complex \textit{reasoning} to balance the object's affordance vs. its functionality. For example, the label \textit{peanut butter} was not in the top-5 labels for the noun initially. Still, the inference process considered it as a possible object based on the presence of the verb \textit{spread} and helped arrive at the final interpretation, which captures the semantics of the scene beyond semantic correspondences between nouns.
\section{Discussion and Future Work}
In this paper, we presented one of the first works on open-world action recognition in egocentric videos. Furthermore, we demonstrated that commonsense knowledge could help break the ever-increasing demands on training data quality and quantity. 
We show that with an initial, trained vocabulary of object (noun) concepts, we can significantly expand our vocabulary to encompass domains and even tasks to learn novel concepts grounded in commonsense knowledge. While we demonstrate open-world inference on egocentric videos, we aim to integrate advances in attention-based mechanisms and relational learning approaches to generalize to videos beyond egocentric.
Extensive experiments demonstrate the applicability of the approach to different domains and its highly competitive performance. 
\section*{Acknowledgments}
This research was supported in part by the US National Science Foundation grant IIS 1955230.
{\small
\bibliography{egbib}
}
\end{document}